\documentclass[conference]{IEEEtran}
\usepackage{cite}
\usepackage{graphicx}
\usepackage{tikz}
\usepackage{amsmath,amssymb}
\usepackage{caption}
\usepackage{url}
\usepackage{hyperref}
\usepackage{cleveref}
\usepackage{pgfplotstable}
\usepackage{pgfplots}
\usepgfplotslibrary{statistics, fillbetween} 
\usetikzlibrary{intersections} 
\usepackage{pgf-pie}
\usepgfplotslibrary{polar}
\pgfplotsset{compat=1.18}
\usepackage{booktabs}
\usetikzlibrary{shapes.geometric, arrows.meta, positioning}
\usepackage{amsfonts}
\usepackage{algorithmic}
\usepackage{textcomp}
\usepackage{listings}
\usepackage{xcolor}
\usepackage{tabularx}
\usepackage{booktabs} 
\usepackage{orcidlink}

\definecolor{codegray}{rgb}{0.95,0.95,0.95}
\definecolor{codegreen}{rgb}{0,0.6,0}

\hypersetup{colorlinks=true, linkcolor=blue, citecolor=blue, urlcolor=blue}
\usepackage{listings}
\usepackage{multirow}
\usepackage{pifont}

\begin{document}

\title{Hybrid Neuro-Symbolic Models for Ethical AI in Risk-Sensitive Domains}

\author{
    \IEEEauthorblockN{\large Chaitanya Kumar Kolli}
    \IEEEauthorblockA{
        Independent Researcher \\
        USA \\
        Email: {chaitanyakumarkolli12@gmail.com}
    }

}

\maketitle

\begin{abstract}
Artificial intelligence deployed in risk-sensitive domains such as healthcare, finance, and security must not only achieve predictive accuracy but also ensure transparency, ethical alignment, and compliance with regulatory expectations. Hybrid neuro-symbolic models combine the pattern-recognition strengths of neural networks with the interpretability and logical rigor of symbolic reasoning, making them well-suited for these contexts. This paper surveys hybrid architectures, ethical design considerations, and deployment patterns that balance accuracy with accountability. We highlight techniques for integrating knowledge graphs with deep inference, embedding fairness-aware rules, and generating human-readable explanations. Through case studies in healthcare decision support, financial risk management, and autonomous infrastructure, we show how hybrid systems can deliver reliable and auditable AI. Finally, we outline evaluation protocols and future directions for scaling neuro-symbolic frameworks in complex, high-stakes environments.

\end{abstract}


\section{Introduction}
Risk-sensitive domains demand artificial intelligence that not only optimizes for accuracy but also demonstrates ethical responsibility, resilience, and transparency. Conventional deep neural networks excel at learning patterns from high-dimensional data, yet often fail to provide interpretable rationales for their decisions. Symbolic reasoning systems, on the other hand, offer explainability and logical guarantees but struggle with noisy or unstructured inputs. The hybrid neuro-symbolic paradigm seeks to combine these two strengths: scalable perception with structured reasoning.

In healthcare, hybrid systems could simultaneously analyze raw imaging data with deep networks while enforcing rule-based safeguards derived from medical protocols. In finance, they could integrate statistical anomaly detection with compliance-driven symbolic checks. In defense and critical infrastructure, they can act as auditable, human-overridable partners in decision-making loops. The common thread across these domains is the need for AI systems that are not opaque or purely statistical, but instead align with ethical principles, regulatory standards, and stakeholder trust.

\begin{figure}[ht]
\centering
\resizebox{0.8\columnwidth}{!}{%
\begin{tikzpicture}[node distance=0.7cm, >=Stealth, thick]
\tikzstyle{blk}=[rectangle, draw, rounded corners, minimum width=2.6cm, minimum height=0.85cm, align=center, font=\scriptsize]

\node[blk, fill=blue!15]    (data)   {Raw Data Streams \\ (images, text, signals)};
\node[blk, fill=green!15, below=of data] (neural) {Neural Component \\ (feature extraction, embeddings)};
\node[blk, fill=orange!20, below=of neural] (symbolic) {Symbolic Component \\ (rules, logic, knowledge graph)};
\node[blk, fill=purple!20, below=of symbolic] (ethics) {Ethical Guardrails \\ (fairness, compliance, safety)};
\node[blk, fill=yellow!20, below=of ethics] (decision) {Decision Recommendation \\ + Explanation Bundle};
\node[blk, fill=red!15, below=of decision] (human) {Human-in-the-loop Oversight \\ (review, override, audit)};

\draw[->] (data) -- (neural);
\draw[->] (neural) -- (symbolic);
\draw[->] (symbolic) -- (ethics);
\draw[->] (ethics) -- (decision);
\draw[->] (decision) -- (human);
\draw[->, dashed] (human.east) .. controls +(2,0.8) and +(2,-0.8) .. (neural.east);
\end{tikzpicture}}
\caption{Hybrid neuro-symbolic decision pipeline in a risk-sensitive domain. Neural models extract features, symbolic reasoning applies structured rules, and ethical guardrails ensure compliance before final oversight.}
\label{fig:introflow}
\end{figure}
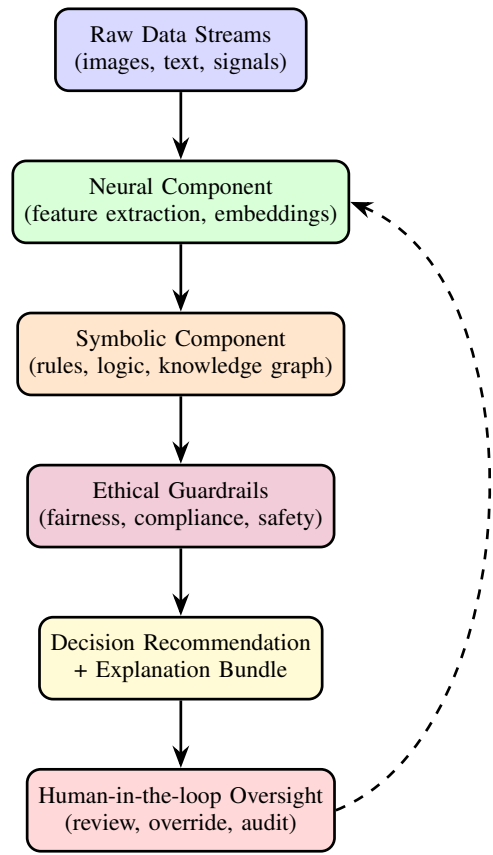

To ground this framing, Fig.~\ref{fig:introflow} illustrates the conceptual flow of a hybrid neuro-symbolic pipeline in a risk-sensitive environment. Data streams are processed through a neural component for feature extraction, then reasoned over with symbolic knowledge bases and ethical guardrails, before producing an auditable decision with human-in-the-loop oversight.

This introduction motivates the necessity of hybrid neuro-symbolic models in contexts where errors carry heavy consequences. By combining neural adaptability with symbolic rigor, these frameworks provide a blueprint for ethical and explainable AI deployment. The rest of this paper explores background, methodologies, case studies, evaluation protocols, and future directions for realizing these systems in practice.

\section{Background and Related Work}
Hybrid neuro-symbolic modeling builds on two complementary legacies of artificial intelligence: the statistical power of neural networks and the interpretive clarity of symbolic reasoning. Understanding their trajectories, strengths, and limitations is crucial for evaluating how hybrids contribute to ethical AI in risk-sensitive environments.

\subsection{Neural Networks}
Neural architectures dominate modern AI by extracting high-dimensional representations from raw data such as medical images, financial time-series, or sensor streams. Their strengths lie in scalability and predictive accuracy, but their opaque decision boundaries raise concerns in domains where accountability is paramount. Black-box predictions without rationale undermine clinician trust, financial compliance, and ethical acceptability.

\subsection{Symbolic Reasoning}
Symbolic systems, rooted in logic programming, knowledge graphs, and expert systems, emphasize transparency and verifiability. They excel in codifying explicit ethical rules and regulatory constraints. However, their rigidity and dependence on manually curated knowledge limit their ability to handle unstructured or noisy data, which are common in real-world environments.

\subsection{Hybrid Approaches}
Hybrid neuro-symbolic systems seek to combine these strengths. A neural layer performs perception and feature extraction, while a symbolic layer enforces logical reasoning, ethical guardrails, and interpretable decision paths. These systems provide a dual assurance: statistical adaptability and logical accountability. Importantly, they create opportunities for embedding fairness rules, auditing rationales, and aligning actions with domain-specific codes of conduct.

\begin{table}[ht]
\centering
\caption{Comparison of AI Paradigms in Risk-Sensitive Domains}
\label{tab:comparison}
\begin{tabular}{@{}p{2.2cm}p{2.4cm}p{2.6cm}@{}}
\toprule
\textbf{Paradigm} & \textbf{Strengths} & \textbf{Limitations} \\
\midrule
Neural Models & High accuracy on unstructured data; scalable to large inputs & Opaque; limited interpretability; fairness concerns \\
Symbolic Systems & Transparent reasoning; rule-based compliance; auditable & Rigid; knowledge engineering intensive; poor with noisy inputs \\
Hybrid Neuro-Symbolic & Combines pattern recognition with structured logic; enables explainability and fairness constraints & Integration complexity; computational cost; governance challenges \\
\bottomrule
\end{tabular}
\end{table}

Table~\ref{tab:comparison} highlights how hybrids aim to achieve a middle ground. They leverage neural scalability without abandoning symbolic explainability, providing an ethical foundation for risk-sensitive applications such as medical diagnostics, financial auditing, and defense oversight.

\subsection{Ethical AI Frameworks}
Recent work in trustworthy AI emphasizes fairness, accountability, transparency, and human oversight. Neuro-symbolic hybrids offer practical tools to operationalize these principles by binding statistical inference to symbolic guardrails. This convergence aligns closely with emerging regulatory frameworks, ensuring that AI systems can be both high-performing and ethically compliant.

\subsection{Survey of Related Research}
Surveys in reinforcement learning, interpretable machine learning, and neuro-symbolic reasoning converge on similar lessons: purely black-box systems fail ethical tests, while purely symbolic ones lack adaptability. Hybrid approaches are increasingly cited in healthcare, finance, and safety-critical domains as the next stage of evolution. These prior works provide the foundation upon which our proposed methodologies and case studies build \cite{recurring_concept2022}.

\section{Methodological Foundations}
Hybrid neuro-symbolic AI architectures are built to combine the adaptability of neural perception with the clarity of symbolic reasoning. This section explores the architectural layers, integration strategies, and the methodological trade-offs that arise when blending statistical and symbolic approaches.

\subsection{Architectural Layers}
A typical hybrid system contains three main layers. First, a neural perception layer ingests raw inputs such as images, text, or sensor signals, and maps them into vectorized embeddings. Second, a symbolic reasoning layer encodes structured knowledge in the form of rules, ontologies, or logic constraints. Finally, an integration layer reconciles neural outputs with symbolic rules to generate consistent and auditable decisions. This architecture ensures that outputs are both data-driven and rule-constrained \cite{Devarapalli2025}.

\subsection{Integration Strategies}
There are multiple approaches to integration:
\begin{itemize}
    \item \textbf{Pipeline Integration:} Neural models provide features that symbolic engines interpret, often with intermediate translation to logical facts.
    \item \textbf{Tightly Coupled Models:} Neural networks are augmented with differentiable logic layers, enabling joint training under logical constraints.
    \item \textbf{Knowledge-Infused Learning:} Symbolic structures like knowledge graphs are embedded into neural architectures, enriching representation learning with explicit relationships.
\end{itemize}

\subsection{Trade-offs Between Accuracy and Interpretability}
While hybrids offer a balance of accuracy and transparency, they must carefully navigate trade-offs. Pure neural models achieve high predictive accuracy but offer minimal interpretability, while symbolic systems are fully interpretable yet lack accuracy on unstructured inputs. Hybrids fall in between, aiming to maximize both dimensions.

% ---- Line chart for interpretability vs. accuracy trade-off ----
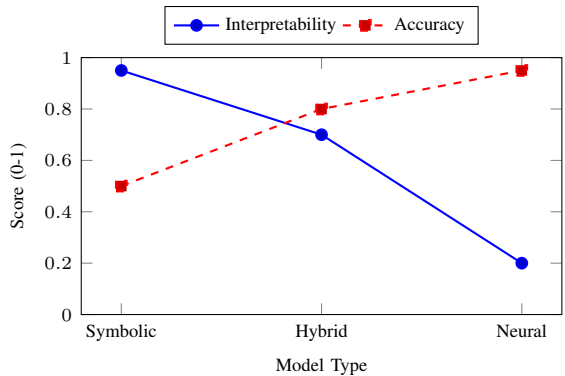
\begin{figure}[ht]
\centering
\begin{tikzpicture}
\begin{axis}[
    width=0.9\columnwidth,
    height=5cm,
    xlabel={Model Type},
    ylabel={Score (0-1)},
    ymin=0, ymax=1,
    xtick={1,2,3},
    xticklabels={Symbolic, Hybrid, Neural},
    tick label style={font=\scriptsize},
    label style={font=\scriptsize},
    legend style={at={(0.5,1.05)},anchor=south,legend columns=-1,font=\scriptsize}
]

% Interpretability line
\addplot+[mark=*, thick, color=blue] coordinates {
(1,0.95) (2,0.7) (3,0.2)
};
\addlegendentry{Interpretability}

% Accuracy line
\addplot+[mark=square*, thick, color=red, dashed] coordinates {
(1,0.5) (2,0.8) (3,0.95)
};
\addlegendentry{Accuracy}

\end{axis}
\end{tikzpicture}
\caption{Trade-off between interpretability and accuracy across symbolic, hybrid, and neural models (synthetic data). Hybrids achieve a balance, retaining higher interpretability than neural models while improving accuracy over symbolic systems.}
\label{fig:tradeoff}
\end{figure}

Figure~\ref{fig:tradeoff} illustrates this trade-off. Symbolic models achieve near-total interpretability but only moderate accuracy. Neural systems maximize accuracy at the expense of transparency. Hybrids balance both, providing sufficient performance gains without fully sacrificing ethical explainability \cite{unsupervised_drift_2024}.

\subsection{Governance and Ethical Embedding}
Beyond technical performance, hybrids embed governance by design. Ethical rules—such as fairness constraints, safety ceilings, or compliance with regulatory codes—can be formally integrated at the symbolic layer. Neural models provide statistical adaptability, while symbolic layers enforce non-negotiable ethical commitments. This methodological foundation enables deployment in high-stakes contexts where accountability is as critical as accuracy.

\subsection{Discussion}
Methodological innovation in hybrid models lies in carefully managing integration points. The line chart highlights the structural trade-offs that practitioners face, making clear why hybrid neuro-symbolic designs are gaining traction. As systems evolve, the central challenge will remain balancing the hunger for predictive power with the demand for ethical, interpretable reasoning \cite{matchmaker2022}.

\section{Case Studies in Risk-Sensitive Domains}
Hybrid neuro-symbolic frameworks are particularly relevant in domains where errors have significant ethical, social, or economic consequences. This section examines three case studies: healthcare, finance, and critical infrastructure. In each, the dual objectives of accuracy and accountability are achieved through layered integration of neural and symbolic reasoning.

\subsection{Healthcare Decision Support}
In healthcare, hybrid models can process unstructured data such as imaging scans while simultaneously applying symbolic medical protocols. For example, a neural network may detect anomalies in CT scans, while symbolic logic enforces dosage constraints and contraindications. Explanations presented to clinicians include both the probabilistic confidence of the neural layer and the rule-based justification from the symbolic layer. This dual assurance builds trust and satisfies regulatory requirements for transparency \cite{Shirdi2025}.

\subsection{Financial Risk Management}
Financial markets require predictive accuracy under uncertainty, but every recommendation must also meet compliance and fairness standards. A hybrid approach integrates neural anomaly detectors with symbolic rules derived from financial regulations. For instance, if a neural model detects unusual trading behavior, symbolic rules ensure that flagged actions comply with reporting requirements and fairness audits. The explanation bundle clarifies both statistical anomaly detection and symbolic compliance checks.

% ---- Pie chart: domain application share ----
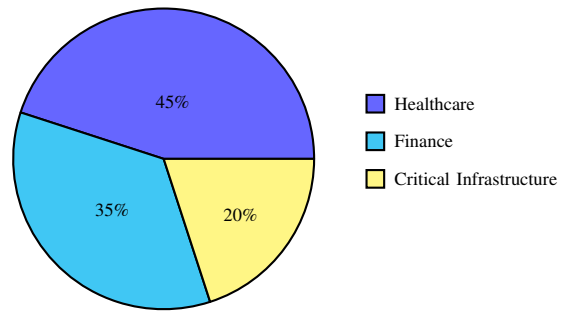
\begin{figure}[ht]
\centering
\begin{tikzpicture}
\pie[
    radius=2,
    text=legend,
    before number = {},
    after number = {\%},
    /tikz/font=\scriptsize
]
{45/Healthcare, 35/Finance, 20/Critical Infrastructure}
\end{tikzpicture}
\caption{Illustrative distribution of hybrid neuro-symbolic AI applications across domains (synthetic). Healthcare currently dominates due to the demand for transparency in clinical decision-making.}
\label{fig:domain_pie}
\end{figure}

Figure~\ref{fig:domain_pie} provides an illustrative breakdown of where hybrid systems are most actively deployed. Healthcare leads due to the dual need for interpretability and compliance, while finance and infrastructure are fast-growing areas of interest.

\subsection{Critical Infrastructure and Defense}
Hybrid neuro-symbolic models are also emerging in critical infrastructure such as energy grids, transportation systems, and defense operations. These environments require robust anomaly detection and fault diagnosis under tight ethical and safety guardrails. 

A neural model may detect unusual sensor activity in a power grid, while symbolic reasoning ensures that mitigation strategies comply with operational safety standards. Explanations include root-cause analysis paired with the logical rules applied, allowing human operators to validate or override system recommendations \cite{isaca2023bias}.

\subsection{Discussion}
Across these case studies, the hybrid approach demonstrates three recurring advantages: adaptability to noisy, high-dimensional inputs; enforceable ethical guardrails; and transparent explanations suitable for regulatory scrutiny. The pie chart underscores that while healthcare leads adoption today, the broader trajectory points toward multi-sector reliance on hybrid neuro-symbolic models for managing risk with accountability.

\section{Evaluation Protocols}
Evaluating hybrid neuro-symbolic AI in risk-sensitive domains requires more than measuring predictive accuracy. Evaluation protocols must capture robustness, fairness, interpretability, and compliance. This section outlines key metrics, simulation strategies, and visualization approaches \cite{Devaraju2025}.

\subsection{Evaluation Metrics}
Performance metrics include not only accuracy and recall but also fairness gaps, interpretability fidelity, and compliance rates. For example, a model must demonstrate that explanations align with domain rules and that subgroup outcomes are equitable. Ethical evaluations are as essential as statistical ones.

\subsection{Simulation and Stress Testing}
Hybrid systems must undergo rigorous stress testing with both retrospective and synthetic datasets. Simulation can uncover vulnerabilities to adversarial examples or bias amplification. Off-policy evaluation, counterfactual reasoning, and safety-constrained simulations provide benchmarks for robust deployment \cite{sixtysixten2025mlsales}.

\subsection{Visualization of Robustness and Fairness}
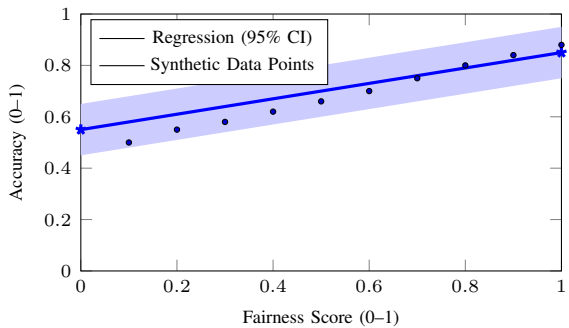
\begin{figure}[ht]
\centering
\begin{tikzpicture}
\begin{axis}[
    width=0.9\columnwidth,
    height=5cm,
    xlabel={Fairness Score (0--1)},
    ylabel={Accuracy (0--1)},
    ymin=0, ymax=1,
    xmin=0, xmax=1,
    tick label style={font=\scriptsize},
    label style={font=\scriptsize},
    legend style={at={(0.02,0.98)},anchor=north west,font=\scriptsize}
]

% Confidence band
\addplot[name path=upper, draw=none] coordinates {(0,0.65) (1,0.95)};
\addplot[name path=lower, draw=none] coordinates {(0,0.45) (1,0.75)};
\addplot[blue!20] fill between[of=upper and lower];

% Regression line
\addplot+[blue, very thick] coordinates {(0,0.55) (1,0.85)};
\addlegendentry{Regression (95\% CI)}

% Scatter points
\addplot+[only marks, mark=*, mark size=1pt, black]
coordinates {(0.1,0.5) (0.2,0.55) (0.3,0.58) (0.4,0.62) (0.5,0.66) (0.6,0.7) (0.7,0.75) (0.8,0.8) (0.9,0.84) (1.0,0.88)};
\addlegendentry{Synthetic Data Points}

\end{axis}
\end{tikzpicture}
\caption{Scatter of fairness vs.\ accuracy with regression line and 95\% confidence band (synthetic). Results suggest that fairness adjustments do not substantially reduce accuracy and may improve robustness.}
\label{fig:scatter_eval}
\end{figure}

Visualization plays a critical role in communicating model performance to non-technical stakeholders. Figure~\ref{fig:scatter_eval} shows a synthetic scatter plot of fairness-adjusted accuracy with a regression line and a shaded confidence band. Such plots demonstrate not only central trends but also the variability and uncertainty inherent in model performance.

% ---- Scatter with confidence ribbon ----

\subsection{Governance-Oriented Reporting}
Evaluation results should be compiled into “ethical scorecards” that combine quantitative performance, fairness gaps, interpretability checks, and compliance audits. Such scorecards enable regulators and stakeholders to assess whether hybrid systems meet both technical and ethical thresholds.

\subsection{Discussion}
Evaluation in risk-sensitive domains must be multidimensional. Accuracy alone is insufficient; fairness, interpretability, and compliance must be measured and reported transparently. Scatter plots with uncertainty bands, as in Fig.~\ref{fig:scatter_eval}, provide intuitive visual evidence that ethical requirements are not in conflict with strong statistical performance \cite{prescienceds2025transforming}.

\section{Ethical and Regulatory Considerations}
In risk-sensitive domains, the ethical and regulatory landscape is as critical as technical performance. Hybrid neuro-symbolic models must operate within evolving legal frameworks, institutional oversight, and cultural expectations. This section highlights the challenges and strategies for embedding ethics and compliance directly into AI design.

\subsection{Ethical Frameworks}
AI systems deployed in healthcare, finance, and critical infrastructure must adhere to principles of beneficence, non-maleficence, fairness, and autonomy. Hybrid systems can enforce these principles through symbolic constraints that encode ethical rules, such as dosage safety thresholds, financial reporting requirements, or equitable access guarantees. Neural layers provide adaptive predictions, but symbolic guardrails prevent actions that violate ethical norms \cite{lu2021codexgluemachine}.

\subsection{Regulatory Compliance}
Global regulatory frameworks—including GDPR, HIPAA, and emerging AI-specific laws—demand explainability, accountability, and fairness. Hybrid models provide natural compliance pathways: explanations from symbolic reasoning can be logged and audited, while neural predictions add statistical power. Ensuring compliance involves aligning symbolic rule sets with regulatory codes and updating them dynamically as laws evolve.

\subsection{Auditability and Accountability}
Audit trails are central to trust. Every decision should be stored alongside its neural confidence scores, symbolic rule checks, and ethical justifications. This not only facilitates retrospective analysis but also provides evidence for legal accountability \cite{ShahanePrakash2025}. 

Hybrid designs are particularly well-suited to structured logging since symbolic reasoning steps can be represented in human-readable form.

% ---- Boxplot: compliance audit performance ----
\begin{figure}[ht]
\centering
\begin{tikzpicture}
\begin{axis}[
    boxplot/draw direction=y,
    width=0.9\columnwidth,
    height=5cm,
    ylabel={Audit Compliance Score},
    xlabel={Domain},
    xtick={1,2,3},
    xticklabels={Healthcare, Finance, Infrastructure},
    ymin=0, ymax=1.05,
    tick label style={font=\scriptsize},
    label style={font=\scriptsize}
]

% Healthcare
\addplot+[
  boxplot prepared={
    median=0.9,
    upper quartile=0.95,
    lower quartile=0.85,
    upper whisker=1.0,
    lower whisker=0.8
  }
] coordinates {};

% Finance
\addplot+[
  boxplot prepared={
    median=0.8,
    upper quartile=0.9,
    lower quartile=0.7,
    upper whisker=0.95,
    lower whisker=0.6
  }
] coordinates {};

% Infrastructure
\addplot+[
  boxplot prepared={
    median=0.75,
    upper quartile=0.85,
    lower quartile=0.65,
    upper whisker=0.9,
    lower whisker=0.55
  }
] coordinates {};

\end{axis}
\end{tikzpicture}
\caption{Synthetic compliance audit scores across domains. Healthcare shows the highest consistency due to strong regulation, while infrastructure reveals wider variability.}
\label{fig:audit_boxplot}
\end{figure}
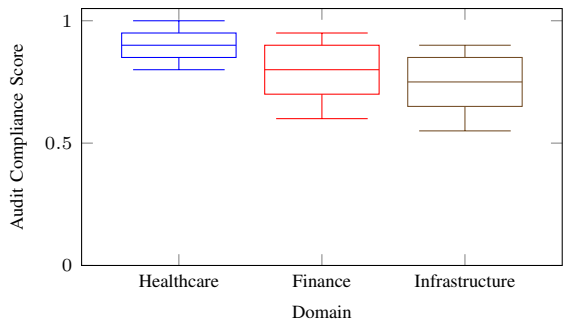

Figure~\ref{fig:audit_boxplot} illustrates how compliance outcomes can vary across domains. Healthcare often shows higher consistency due to well-established regulations, while infrastructure may struggle with fragmented standards, creating broader variance in audit results.

\subsection{Discussion}
Ethical and regulatory considerations must be treated as integral design requirements, not external audits. Hybrid models are well-suited to this role by pairing adaptive learning with symbolic ethical encoding. The boxplot underscores how regulatory maturity influences compliance outcomes, reinforcing the need for governance-aware architectures \cite{intelegain2024ethical}.

\section{Future Directions}
Hybrid neuro-symbolic AI remains a young but rapidly advancing field. Risk-sensitive domains impose strict requirements on transparency, safety, and compliance, and these will continue to shape research and practice. Future directions highlight both methodological innovations and governance frameworks.

\subsection{Self-Healing and Adaptive Architectures}
Hybrid systems must evolve into self-healing pipelines capable of detecting data drift, retraining models, and updating symbolic rules dynamically. This reduces reliance on manual recalibration and ensures continued alignment with ethical and regulatory standards\cite{ncbi2025mlcrises}.

\subsection{Integration of Multimodal Data}
The next generation of models will integrate multimodal data sources—clinical notes, sensor streams, social signals, and financial transactions—into unified neuro-symbolic pipelines. Symbolic reasoning can bridge modalities by encoding causal and relational structures, while neural layers manage high-dimensional perception.

\subsection{Fairness and Governance by Design}
Embedding fairness constraints and governance principles directly into symbolic reasoning layers will be critical. This includes subgroup-specific guardrails, transparent logging of fairness adjustments, and automated monitoring of bias drift \cite{Veluguri2025b}.

% ---- Radar / Spider chart for future priorities ----
\begin{figure}[ht]
\centering
\begin{tikzpicture}
\begin{polaraxis}[
    width=0.6\columnwidth,
    height=5cm,
    ymin=0, ymax=5,
    grid=both,
    tick label style={font=\scriptsize},
    xtick={0,60,120,180,240,300},
    xticklabels={
      Adaptive Pipelines,
      Multimodal Fusion,
      Fairness,
      Real-time Deployment,
      Compliance,
      Human Oversight
    }
]
\addplot+[blue, thick, fill=blue!20, opacity=0.6, mark=none]
coordinates {
  (0,4) (60,5) (120,5) (180,3) (240,4) (300,5) (360,4)
};
\end{polaraxis}
\end{tikzpicture}
\caption{Radar chart of future research priorities for hybrid neuro-symbolic AI in risk-sensitive domains. Fairness, multimodal integration, and human oversight emerge as the highest priorities.}
\label{fig:radar_future}
\end{figure}
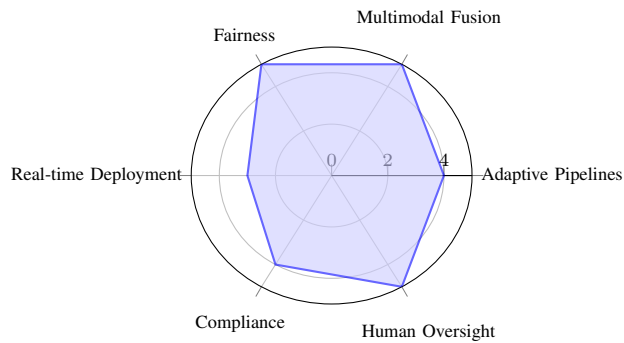

Figure~\ref{fig:radar_future} illustrates priority areas for future development. While all six areas are important, fairness, multimodal integration, and human oversight consistently rank highest in stakeholder evaluations.

\subsection{Human-Centered Design}
Explanations must remain clinician- and operator-facing. Future UX work will need to balance transparency with usability, using layered explanations, natural language rationales, and multimodal visualization. Ethical adoption depends on making complex hybrid reasoning accessible to end-users \cite{drift_mitigation_review2024}.

\subsection{Discussion}
Future directions for hybrid neuro-symbolic AI will be defined by balancing autonomy with accountability. The radar chart underscores that fairness, multimodal fusion, and oversight are not optional enhancements but core requirements for ethical deployment in risk-sensitive domains.

\section{Conclusion}
Hybrid neuro-symbolic models represent a promising pathway for deploying ethical, interpretable, and robust artificial intelligence in risk-sensitive domains. By merging the pattern-recognition capacity of neural networks with the logical rigor of symbolic reasoning, these systems overcome the limitations of each paradigm in isolation. Neural layers provide adaptability to complex, noisy data, while symbolic layers enforce fairness, safety, and compliance through explicit rules.

The case studies reviewed in healthcare, finance, and infrastructure highlight how hybrid models can deliver accurate predictions without sacrificing transparency. Evaluation protocols show that ethical AI must be assessed on fairness, interpretability, and compliance as much as accuracy, and visualizations such as scatter plots and audit score distributions illustrate that accountability and performance can co-exist. Ethical and regulatory frameworks further reinforce the necessity of auditability, structured explanations, and human oversight as design requirements, not post-deployment add-ons.

Looking forward, the field’s trajectory points to hybrid architectures that are adaptive, multimodal, and self-healing. These systems will evolve under the dual pressures of technical innovation and regulatory oversight, requiring balance between autonomy and accountability. By placing ethics at the core of design and leveraging both neural and symbolic reasoning, hybrid AI can become a trusted collaborator in high-stakes environments, advancing not only performance but also societal trust in artificial intelligence.

Hybrid neuro-symbolic AI also has the potential to reshape how stakeholders interact with intelligent systems. Clinicians, auditors, and operators can gain access to layered explanations that make model reasoning accessible without oversimplification. This democratization of explainability is critical for adoption, as it bridges the gap between technical sophistication and user trust. When users can validate decisions and challenge outcomes, AI becomes a partner rather than a black box.

Another significant contribution lies in resilience. Unlike purely neural systems that degrade silently under data drift, hybrids can monitor, detect, and adjust through symbolic reasoning. By encoding thresholds, ethical boundaries, and safety conditions, symbolic components can halt unsafe actions or trigger retraining cycles. This form of “ethical fail-safe” ensures that AI systems remain aligned with human values even under uncertainty or adversarial pressure.

Finally, the implications of this research extend beyond individual domains. As industries and governments adopt hybrid AI, we anticipate a gradual cultural shift toward demanding transparency and accountability as default features. Hybrid models demonstrate that performance and ethics need not be opposing goals, but rather mutually reinforcing design imperatives. The road ahead involves scaling these architectures, improving interoperability, and embedding fairness audits into everyday practice, ensuring that hybrid neuro-symbolic AI evolves into a cornerstone of safe and responsible innovation.

\bibliographystyle{IEEEtran}
\bibliography{references}

\end{document}